\documentclass{IEEEtran}
\usepackage{amsmath}
\usepackage{amssymb}
\usepackage{color}
\usepackage{natbib}
\usepackage{graphicx}
\usepackage{flushend}
\title{Fast nonparametric clustering\\ of structured time-series}
\author{James Hensman, Magnus Rattray and Neil D. Lawrence
\IEEEcompsocitemizethanks{\IEEEcompsocthanksitem J. Hensman and N.D. Lawrence are with the Department of Computer Science and Sheffield Institute for Translational Neuroscience, University of Sheffield, UK\protect\\
	M. Rattray is with the Faculty of Life Science, University of Manchester, UK
}
}

\newcommand{\bY}{\mathbf Y}
\newcommand{\by}{\mathbf y}
\newcommand{\bg}{\mathbf g}
\newcommand{\bt}{\mathbf t}
\newcommand{\bff}{\mathbf f}
\newcommand{\bT}{\mathbf T}
\newcommand{\bK}{\mathbf K}
\newcommand{\bI}{\mathbf I}
\newcommand{\gp}{\mathcal {GP} }
\newcommand{\bbE}{\mathbb E}
\newcommand{\sumover}[1]{\sum_{#1=1}^{\MakeUppercase #1}}
\newcommand{\prodover}[1]{\prod_{#1=1}^{\MakeUppercase #1}}

\newcommand{\ktilde}{\widetilde k }
\newcommand{\Ktilde}{{\bf \widetilde K }}
\newcommand{\yhat}{{\bf \hat y}}
\newcommand{\that}{{\bf \hat t}}
\newcommand{\nhat}{{\bf \hat n}}
\newcommand{\given}{\,|\,}
\newcommand{\normal}{\mathcal N}
\newcommand{\btheta}{\boldsymbol \theta}
\newcommand{\bgamma}{\boldsymbol \gamma}
\newcommand{\bphi}{\boldsymbol \phi}
\newcommand{\bzero}{{\bf 0}}
\newcommand{\bZ}{{\bf Z}}
\newcommand{\bV}{{\bf V}}
\newcommand{\bF}{{\bf F}}
\newcommand{\bz}{{\bf z}}
\newcommand{\bbR}{\mathbb R}
\renewcommand{\d}{\,\text{d}}
\renewcommand{\th}{^\text{th}}
\newcommand{\correction}[1]{ #1}

\begin{document} 
\maketitle 
\IEEEcompsoctitleabstractindextext{%
\begin{abstract}
In this publication, we combine two Bayesian nonparametric models: the Gaussian
Process (GP) and the Dirichlet Process (DP). Our innovation in the GP model is
to introduce a variation on the GP prior which enables us to model {\em
structured} time-series data, i.e. data containing groups where we wish to
model inter- and intra-group variability.  Our innovation in the DP model is an
implementation of a new fast collapsed variational inference procedure which
enables us to optimize our variational approximation significantly faster than
standard VB approaches. In a biological time series application we show how our
model better captures salient features of the data, leading to better
consistency with existing biological classifications, while the associated
inference algorithm provides a significant speed-up over EM-based variational inference. 
\end{abstract}
}

\maketitle

\IEEEdisplaynotcompsoctitleabstractindextext
\section{Introduction}

We consider the problem of modelling and clustering structured time series. We
turn to two tools from the field of Bayesian nonparametrics, using Gaussian
Processes (GPs) to model time series and Dirichlet Processes to model clusterings. 

\correction{Our model is constructed as follows. Given some data which is partitioned
into disjoint groups (or batches), we construct a hierarchical GP model, using
a GP to model each group, and a single additional GP to model the prior mean
for the whole set.}  The {\em general behaviour} of the groups is governed by
the last GP, and the deviations of each group from this mean behaviour is also
modelled as a GP. We envisage that for many applications, these groupings may
have sub-divisions, which we model using additional layers of hierarchy.
Further, we construct a model where the top level partition is unknown
a-priori, i.e. in the case of clustering, using a Dirichlet Process (DP) prior
with a GP base distribution, each atom of which becomes the prior mean for a
hierarchical GP. This allows us to perform inference over clusters of
hierarchically grouped data. 

Our model is inspired by the analysis of gene-expression time-series data.
Previous models for clustering time-series using GP models
\citep{dunson2010nonparametric, cooke2011bayesian} failed to account for
structure in the data and \correction{previously proposed inference
procedures (Gibbs sampling, agglomerative clustering) do not scale well.}
In this biological application, values of gene expression are measured at
regular or irregular time intervals spanning some phenomenon such as
development or disease progression. The high cost of measurement or
limited temporal resolution of the underlying system usually dictates a
small number of time points, and the measurement process is subject to
both technical and biological variation.  Groups in the data occur
naturally: time series may be taken for different patients in a clinical
trial, thus grouping them by patient number; measurements can be taken
during development of related species  or subject to replicated
experiments. We can envisage more than one level of grouping: we might
have different patients with measurements taken at different hospitals,
say, or developmental time series taken by different laboratories using
different technologies. 


Whilst inference in a GP is tractable\footnote{subject to known or optimised
hyper-parameters and a Gaussian likelihood}, inference in a DP requires some
numerical procedure such as Gibbs sampling or variational approaches. Whilst
variational methods are widely acknowledged to be faster than sampling, there
is a need for {\em even faster} inference in these methods. In particular,
faster inference allows for the exploration of larger datasets given the same
computational resources and avoids the common practice of applying a crude
filtering to reduce the size of data set prior to modelling. We propose a fast
inference scheme based on recent work by \cite{hensman2012fast}.

This novel derivation of variational Bayes (VB) amalgamates several key ideas
from the literature.  First, we construct a {\em KL-corrected}
\citep{king2006fast} bound on the marginal likelihood, where our objective
function depends only on the approximate distribution of the clustering
(latent) variables.  The other model parameters are marginalised after
constructing a lower bound on the {\em conditional} likelihood, and are not
explicitly parameterised in optimization.  This makes the parameter-space of
the optimization significantly smaller. On this reduced parameter space, we
make use of the Riemann structure of the approximation to derive the {\em
natural} gradient, which is closely linked to the VBEM procedure. Using
approximate geometrical conjugacy on the manifold
\citep{honkela2010approximate}, we implement a conjugate natural gradient
approach that outperforms VBEM and free-form optimization. 


\section{Related work}
Gaussian Process methods have been applied to gene expression time-series with several aims, such as to infer transcription regulation \citep{honkela2010model}, and to find dynamic differential expression \citep{kalaitzis2011simple, stegle2010robust}. Recently, we have proposed hierarchical Gaussian processes \citep{hensman2012hierarchical} for modelling gene expression, and showed that this simple proposal led to much improved modelling of the time-series with various applications. 
\correction{
\citet{park2010hierarchical} also proposed a method based on hierarchical Gaussian processes. Their presentation is conceptually similar, but with the objective of saving some computation in performing Gaussian process regression, and \citet{behseta2005hierarchical} proposed a hierarchical Gaussian process model with application to neuronal Poisson-process intensities. }

Clustering gene expression time series is an application which has attracted a lot of interest. Analysis of time series clusters is an important tool in exploring and understanding gene networks,  whilst incorporating knowledge of the time-series into the model has the potential to improve the ability of the method to discern clusters. \citet{dunson2010nonparametric} proposed a DP-GP model much like ours, but we make some important additions. In Dunson's model, a series of GP functions are drawn from a DP-GP prior, and each observation is then assigned to one of the functions. However Dunson makes no use of {\em structure} in the model: observations differ from the latent function draws only by white noise. In our model, we use further GPs to describe how genes differ from the latent function, and {\em further} GPs to describe more structure in the experiment, such as replications. \citet{rasmussen2002infinite} Also presented a method which combined GPs using DPs, but this method used a {\em gating} approach to produce a mixture of experts model, with different aims to that presented here. 

Outside the nonparametric framework, \citet{medvedovic2004bayesian} has described a hierarchical clustering model, but our approach is novel in applying structure to a GP model of the time series within the clustering. \citet{cooke2011bayesian} proposed a GP based clustering approach for gene expression, where replicates were used {\em before} clustering to estimate the level of noise in the experiment. Our richer model explicitly accounts for replicate structure in the experiment. 

Our method for inferring structure and clustering in the model also offers an improvement over the aforementioned approaches. We show that our inference method is considerably faster than the usual variational method (VBEM), which is widely acknowledged to be faster than the  Gibbs sampling approach adopted by \citet{dunson2010nonparametric}. The agglomerative clustering method proposed by \citet{cooke2011bayesian} will suffer significant scalability issues: in the first round of agglomeration one needs to compute marginal likelihoods for GP models of {\em every} pair of genes. We shall derive a collapsed variational procedure for inference of clustering. 

Collapsed variational Bayes (CVB) \citep{kurihara2007collapsed, teh2007collapsed}, and the latent variable method of \citet{sung2008latent} share many properties of our approach, see \citep{hensman2012fast} for details. Yet our method improves on CVB by considering the Riemannian structure of the collapsed problem, and relating it to the VBEM method. We can derive extremely efficient gradient methods, and apply conjugate gradient algorithms which account for the Riemman structure of the collapsed problem. Previously, \citet{honkela2010approximate} proposed natural gradient based variational methods, but without connection to the collapsed approach they were unable to achieve a speed-up over standard VBEM. 


A related model which uses the KL-corrected approach is {\em Overlapping mixtures of Gaussian Processes} \citep{lazaro2011overlapping}. In this model, a series of latent GP functions are assumed, to which each observation is then assigned, with the objective of tracking. Our model can be reduced to this by removing the structured GP element, as well as the DP prior. Further \citep{lazaro2011overlapping} used free-form variational optimization, which we shall show to be much slower than our approach.

\section{Hierarchical Gaussian Processes}

In this section, we briefly review Gaussian Process regression and introduce
our notation. We then extend the GPs to model structured time series before
introducing our notation for mixture models. 

Gaussian process (GP) regression is perhaps the most widely applied of Bayesian
nonparametric methods, particularly since the publication of
\cite{rasmussen2006gaussian}. The idea is to place a prior over the space of
functions, and use Bayesian inference to update one's belief in the function by
observing noisy realisations of the function at a finite number of points. 

The GP is specified by a mean function $m(t)$ and a covariance function
$k(t,t')$.  The mean function is often assumed to be zero everywhere and
the covariance function takes a parametric form. In this publication we make use of the square-exponential covariance function: $k(t,t') = \sigma^2 \exp\{\tfrac{-1}{2\ell^2} (t-t')^2\}$.
The parameters of the covariance function control the type of function permitted: the variance of the function is controlled by the parameter $\sigma^2$, and the length-scale of the function by $\ell$. In this publication where we may have several separate covariance functions, we use subscripts to identify the function to which the parameters belong, and the vector $\btheta$ to collect appropriate covariance function parameters. The prior over functions is written
\begin{equation}
	p(f) = \gp \left(m(t), k(t,t')\right).  
\end{equation}
The critical property of a GP is that the distribution of any finite set of
function values has a Gaussian distribution. If the vector $\bff$ contains
the values of the function $f$ at times $\bt$, the prior over $\bff$ is
\begin{equation}
p({\bf f}|\bt, \theta) = \normal\left(\bff| \bzero,\,\bK(\bt, \bt)\right),
\end{equation}
where $\bK(\bt,\bt)$ is a covariance matrix constructed from the covariance function: $\bK(\bt,\bt)[i,j] = k(\bt[i],\bt[j])$. 
 In the regression setting, we are usually presented with a noise corrupted version of $\bff$, $\by$. Assuming that the noise is Gaussian, writing
 \begin{equation}
	 p(\by|\bff) = \normal \left(\by\given \bff, \beta \bI\right),
 \end{equation}
it is trivial to marginalise the values $\bff$ due to the conjugate nature of the Gaussian noise and the Gaussian process prior:
 \begin{equation}
	 p(\by|\bt, \btheta) = \normal \left(\by\given \bzero, \bK(\bt,\bt) + \beta \bI\right).  
 \end{equation}

One interpretation of this is that there is a function $y(t)$ with Gaussian process prior covariance $k_y(t,t') = k_f(t,t') + \beta\delta(t,t')$, which we observe directly.  It is this conjugate relationship that we will use to construct structured models, using not i.i.d white noise, but further Gaussian process functions. 

\subsection{GPs for structured time series}
Consider a set of time series which we wish to model. We have $N$ groups of
data $\bY = \{\by_n\}_{n=1}^N$ taken at times $\bT = \{\bt_n\}_{n=1}^N$. For
example, each group could represent an experimental replication under different
conditions. 
Under our model, there is a latent GP function which governs all
the time series, which we denote $f(t) \sim \gp( 0, {k_f(t,t')})$. Given a draw
for $f$, each group of data is then drawn from a GP: $y_n(t) \sim \gp ({f},
{k_n(t,t')})$. The additional covariance $k_n$ can account for both correlated structure in $\by$ and noise.  

\begin{figure*}
	\centering
	\includegraphics[width=\textwidth]{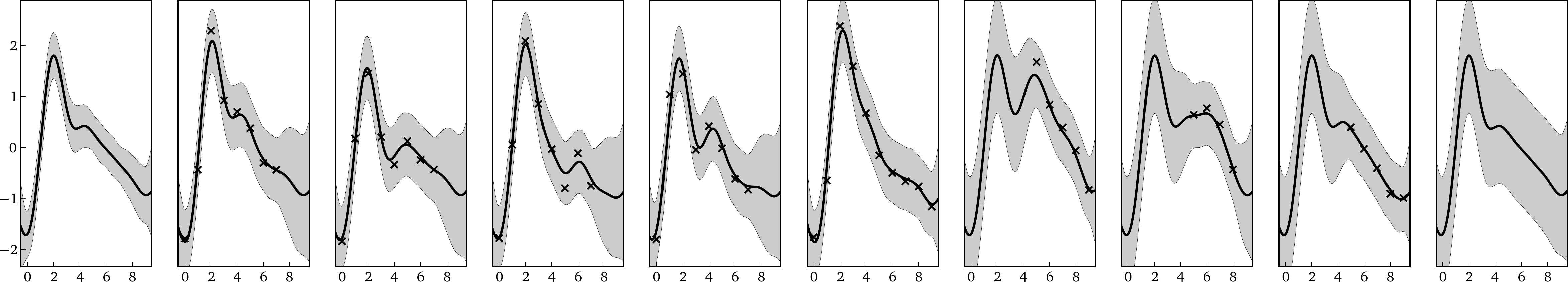}
\caption{A hierarchical Gaussian process model of gene expression of a single gene during {\em Drosophila} development. In each pane, the y-axis represents normalised log gene-expression level, and the x-axis represents time in hours. The left-most frame shows the posterior distribution for the function $f(t)$, and subsequent frames represent biological replicates, which we model as hierarchical groups.  The right-most pane represents a hypothetical future replicate. Posterior means are represented as solid lines and shaded areas represent two standard deviations. \label{fig:hierarchical_illustration}}
\end{figure*}

It is possible to marginalise the latent function and thus introduce covariance
between the groups, using the same conjugate properties as for the noise discussed above. This covariance amongst the groups then depends on the group index $n$, and we can write a compound covariance function
\begin{equation}
	{\ktilde(t,t',n,n')} = \left\{ \begin{array}{rl}
		k_f(t,t') + k_n(t,t') &\mbox{ if $n=n'$} \\
		k_f(t,t') &\mbox{ otherwise. }
\end{array} \right.
\label{eq:compound_covariance}
\end{equation}

Considering the group index $n$ as an {\em input} to the function, we can write
\begin{equation}
	y(t,n) \sim \gp \left(0, {\ktilde(t,t',n,n')}\right).  
\end{equation}
To obtain a likelihood for the grouped data, we concatenate $\bY$ as $\yhat = [\by_1^\top  \ldots \by_N^\top]^\top$  and similarly for $\bT$ (constructing $\that$) and the group indexes ($\nhat$). We construct a kernel matrix $\Ktilde$ on the concatenated vectors $\that, \nhat$ such that $\Ktilde[i,j] = \ktilde(\that[i],\that[j],\nhat[i],\nhat[j])$ and write
\begin{equation}
p(\bY \given \bT,\btheta) = \normal \left(\yhat \given \bzero, \Ktilde\right),
\end{equation}
where $\btheta$ is a vector of all covariance function parameters which we can then infer by type-II maximum likelihood or MCMC sampling. 

Figure \ref{fig:hierarchical_illustration} illustrates a simple application of this idea. This data represents a single gene in the {\em Drosophila} development dataset (see section \ref{par:dros_dev}). 

\section{Mixtures of GPs}
Mixture models involve the assignment of data into clusters, as well as the inference of the properties of each cluster. The popular Gaussian mixture model involves the assignment of each data vector to one of $K$ Gaussian components, whilst simultaneously inferring the mean and covariance of each of the components. The EM algorithm for Gaussian mixture models is widely known: it treats the assignment of data to clusters as a latent variable problem, and estimates the cluster means and variances by maximising the likelihood. The algorithm alternates between inference of the latent variables and maximisation of the likelihood with respect to the parameters. 

In Bayesian mixture models, the cluster parameters are treated as random variables, and inference is performed by computing (or approximating) the joint posterior distribution of the latent variables and cluster parameters.  A variational approach to inference in a mixture model can be achieved by approximating the posterior with a factorising distribution, which is updated using the VBEM algorithm. Using a set of conjugate priors, the VBEM algorithm alternates between computing the optimal approximating distribution for the assignment variables (the VB-E step), and finding that for the cluster parameters and mixing proportions (the VB-M step). The variational procedure is better than the EM method since it avoids the pitfalls of maximum likelihood estimation, however both methods require the specification of the number of  components, $K$. 

Using a Dirichlet process prior for the mixing proportions avoids the problem of selecting the number of components in the model.  It also offers a convenient inference procedure via Gibbs sampling, though in this paper we focus on a variational approach. The Dirichlet process can be seen as an infinite mixture model \citep{rasmussen2000infinite}: a normal mixture model where the \correction{atoms correspond to (parameters of) Gaussian densities, and the} number of clusters has been allowed to tend to infinity. \correction{In the most general case, DPs can be used for infinite mixture models, not simply Gaussian densities}. The mixing proportions of the clusters (and hence the expected number of clusters) are controlled by a concentration parameter $\alpha$.

We propose a Dirichlet Process Gaussian Process (DPGP) mixture model, using the stick breaking construction as follows. Let $\Omega$ be a space of functions mapping $\bbR \to \bbR$, and let $P$ be a DP on that space, with a GP base distribution $H = \gp\left( 0, {k_f(t,t')}\right)$ and concentration $\alpha$. We draw a series of atoms and associated stick-breaking lengths independently such that
\begin{equation}
\begin{split}
	f_i \sim 
	\gp\left( 0, {k_f(t,t')}\right) &,\,\,\,i=1\ldots\infty\\
	v_i \sim  \text{beta}(1,\alpha) &,\,\,\,i=1\ldots\infty.
\end{split}
\label{eq:stick_breakers}
\end{equation}
From the stick breaking weights we define a series of mixing proportions $\pi_i = v_i\prod_{j=1}^{i-1}(1-v_j)$, and the distribution $P$ can be written
\begin{equation}
P = \sum_{i=1}^\infty \pi_i \delta_{f_i},
\end{equation}
thus each atom of our DP is a function drawn from a GP. This construction is similar to that provided by \citet{dunson2010nonparametric}, though we have innovated in using additional structure in the model (which we will show empirically to be very effective), and we also propose a novel inference procedure based on variational Bayes.

To use this construction in clustering functional data, we use the hierarchical
GP developed in the previous section. Each group of data $(\bt_n, \by_n)$ is
then associated with a single atom of the DP by the variables $\bZ =
\{\bz_n\}_{n=1}^N$, and varies from the atom by an independent draw from
another GP. Each atom becomes the mean function in a hierarchical GP
described above.

In our applications, we have further levels of this hierarchy, with unknown
groups (clusters) at the highest level and known groups (replicates) at lower levels. It is simple to
extend the model with a series of levels with known and unknown groupings at
each, depending on the application. 

The generative procedure for our model is then:
\begin{enumerate}
	\item Select a concentration parameter $\alpha$, and GP hyperparameters for the DP-GP construct. 
	\item Draw an infinite series of stick-breaking lengths $v_i$ and associated atomic functions $f_i$, compute the infinite mixing proportions $\pi_i$. 
	\item For each group of data, draw the random variable $\bz_n$, thus selecting the atomic function $f$ associated with the group.
	\item For each subgroup in that group, draw a function for a GP which defines the deviation from the selected atomic function (this may itself be a hierarchical GP over sub-sub-groups). 
	\item Evaluate the functions at a finite set of points $\bt_n$, reporting the values $\by_n$. 
\end{enumerate}



Presented with a set of data $\bY, \bT$, perhaps with some known structure, we are tasked with inferring the unknown groupings (via the variables $\bZ$), the latent functions $f(.)$, the GP parameters $\theta$ and the all the functions $y(.)$ which occur in the data.

\section{Inference}
Variational Bayes is a method for probabilistic inference  where the posterior distribution is approximated using some simpler distribution. The usual assumption is that the posterior factorises in some way which yields a tractable lower bound on the marginal likelihood, which then serves as an objective in optimization. The factorising assumption leads naturally to the VBEM algorithm, where each of the factors is updated in turn. 

Recently, various forms of {\em collapsed} variational Bayes have been proposed for specific models \citep{king2006fast, lazaro2011variational, lazaro2011overlapping, teh2007collapsed, kurihara2007collapsed, sung2008latent}, where some of the variables are marginalised analytically. In \citet{hensman2012fast}, we showed that many of these schemes are equivalent. We also proposed a Riemann optimization scheme similar to \citet{honkela2010approximate}, and showed that VBEM is in fact a {\em steepest ascent} algorithm upon a Riemannian manifold. Thus introducing geometrically conjugate gradient directions serves to increase the speed of convergence. Here we consider the application of these ideas to the DP-GP model. 

For Dirichlet process mixture models as we consider here, \citet{kurihara2007collapsed} considered forms of a collapsed stick-breaking prior. Although their approach differs from our derivation in that they marginalised stick breaking lengths {\em before} making a variational approximation, we will show that we end up with similar expressions. Further, we collapse {\em all} of the parameters from our model aside from the cluster allocation variables. This leads to greater simplification in computing gradients and proposing non-gradient moves in optimization such as merge-split and re-ordering of the clusters. 

The steepest ascent direction on a Riemann manifold is given by the {\em natural} gradient \citep{amari1998natural}. For our model, and for related mixture models, we show that the necessary
information-geometric quantities can be computed in closed form {\em without}
the expensive matrix inverse which hindered previous approaches
\citep[such as][]{honkela2010approximate}. We demonstrate empirically that our algorithm converges faster than VBEM and the free-form approach of \citet{lazaro2011variational}. 

We note that the variables $\bz_n$ are each of infinite dimension with a single unitary element such that $\bz_{nk} \in \{1,0\}, \sum_{k=1}^\infty \bz_{nk}=1$, in our approximate posterior, we shall truncate the number of components, selecting a {\em truncation} parameter as in \citep{blei2006variational, kurihara2007collapsed}. To select this we
adopt a merge-split approach which has been applied before in maximum
likelihood clustering \citep{ueda2000split} and also as a Metropolis-Hastings step in a
collapsed Gibbs sampler \citep{jain2004split}.  

First, we follow the procedure outlined by \citet{hensman2012fast} to derive a collapsed lower bound on the marginal likelihood, which serves as an objective function in optimization. 

\subsection{Model definitions}
\begin{figure}
	\centering
	\includegraphics[width=\linewidth]{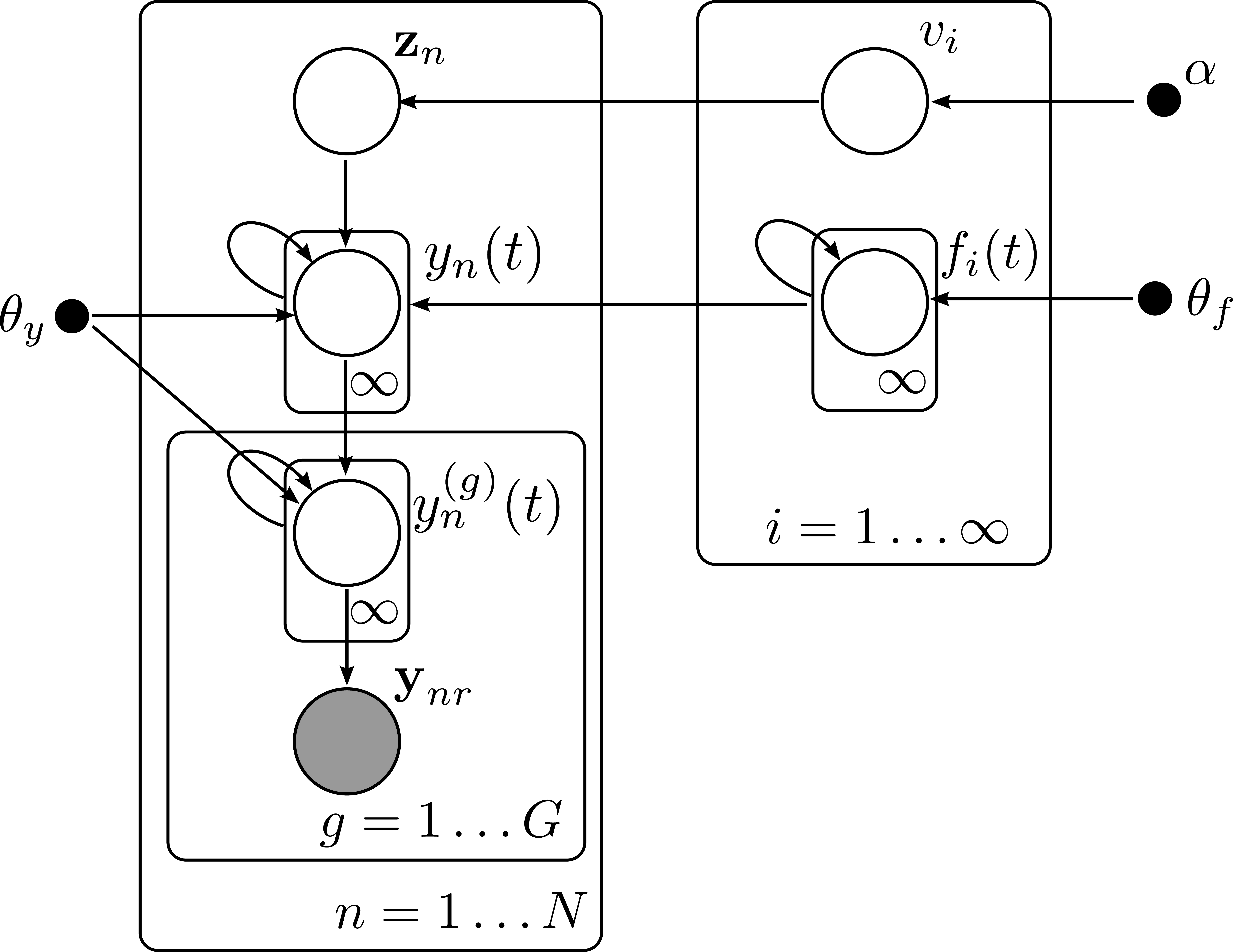}
	\caption{\label{fig:graphical_full} A graphical models representation of our hierarchical Gaussian process clustering method. Gaussian processes are represented by infinite self-connected plates (note that all variables in a GP are jointly distributed).  Hyper-parameters of the GPs and the DP concentration $\alpha$ are shown as solid dots. The right-hand plate represents the Dirichlet process, and the left hand plate represents N independent data groups to be clustered. The inner plate represents a single level of structure below the clustering, indexed by $g$, with functions represented as $y_n^{(g)}(t)$. }
\end{figure}

\begin{figure}
	\centering
	\includegraphics[width=\linewidth]{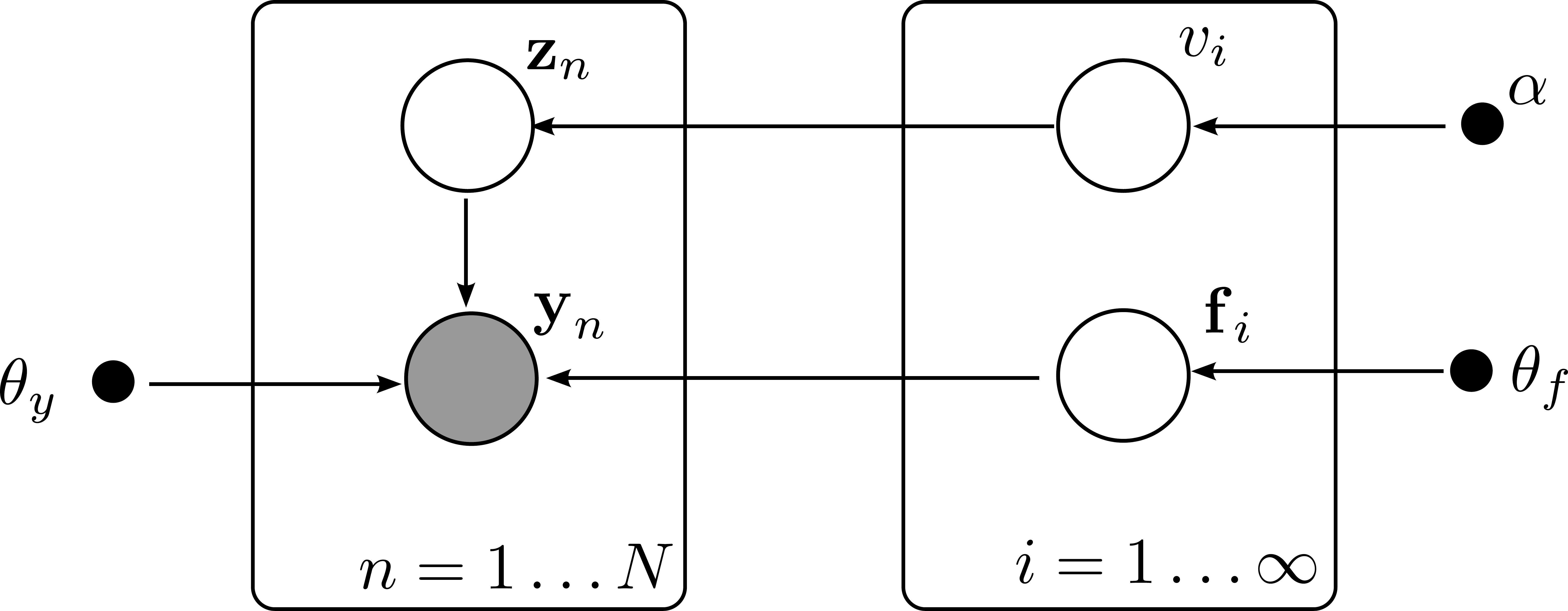}
	\caption{\label{fig:graphical_collapsed} A graphical models representation of our hierarchical Gaussian process clustering method, after variables have been collapsed using the standard GP methodology. This model enables the d-separation test which shows that by approximating the distribution of the  latent variables $\bZ$ with $q(\bZ)$, the remainder of the model will marginalise analytically.}
\end{figure}

We briefly formalise the notations for our model, summarising them in Table \ref{tab:notation}. We have a DP-GP construction as described above, whose atoms we denote $f_i(\cdot)$. Suppose we have $N$ groups of data which we wish to cluster. In gene expression data, the expression of all genes are necessarily gathered at the same time, so each of the vectors $\bt_n$ are the same, simplifying our exposition somewhat. Let the values of the function $f_i$ at times $\bt$ be gathered into the vector $\bff_i$, and denote the collection of these $\bF = \{\bff_i\}_{i=1}^\infty$. Collect the stick breaking lengths similarly $\bV = \{v_i\}_{i=1}^\infty$. From here, the vectors $\bY= \{\by_n\}_{n=1}^N$ represent the data that we wish to cluster, and any sub-groupings have been concatenated. This simplifies the model as illustrated by Figures \ref{fig:graphical_full} and \ref{fig:graphical_collapsed}. Accordingly, we can define the likelihood in the usual mixture form:
\begin{equation}
	p(\bY\given \bF, \bV) = \prod_{n=1}^N\prod_{k=1}^\infty \mathcal N(\by_n\given\bff_k, \bK_y)^{\bz_{nk}}
\end{equation}
where $\bK_y$ is a covariance matrix constructed according to any sub-groups in $\by_n$ as per equation \eqref{eq:compound_covariance}, the prior for $\bV$ is as defined in equation \eqref{eq:stick_breakers}, and the prior for $\bF$ occurs through the usual GP methodology as 
\begin{equation}
	p(\bF) = \prod_{k=1}^\infty \mathcal N(\bff_k\given \bzero,\bK_f).  
\end{equation}
In the above, the GP hyperparameters have been omitted for brevity: in practise we make point estimates for the parameters interleaving gradient-based optimization with the VB procedure. 

\begin{table*}
	\centering
\caption{Notation for variables used in the model \label{tab:notation}}
\begin{tabular}{|c|c|l|}
symbol& type & description \\
	\hline
$ \bz_n$ & $\in \{0,1\}^\infty$ & Allocates the $n\th$ data group $\bY_n$ to a latent function\\
$\bZ$ & $= \{\bz_n\}_{n=1}^N$ & collection of allocation variables\\
$f_i(\cdot)$ & $:\mathbb R \to \mathbb R $&  the $i\th$ latent function\\
$\bff_i$ & $\in \mathbb R^D$ & realisations of the $i\th$ function $f_i(\cdot)$ at the points $\bt$\\
$\bF$ & $=\{\bff_i\}_{i=1}^\infty$ & collection of all realised function values\\
$\by_n$ & $\in \mathbb R^D $& the $n\th$ group of observed data\\
$\bY$ & $=\{\by_n\}_{n=1}^N$ &  collection of all observed data\\
$v_i$ & $\in [0,1]$ & the $i\th$ stick-breaking length in the DP construction\\
$\bV$ & $=\{v_i\}_{i=1}^\infty$ &  collection of all stick-breaking lengths\\
$\pi_i$ & $= v_i\prod_{j=1}^{i-1}(1-v_j)$&  mixing proportions defined by stick breaking construction\\
$\alpha$ & $ \in \mathbb R^+$ & concentration parameter of the stick breaking process\\
$\phi_{nk}$ & $ \in [0,1]$ & (approximate) posterior probability of assigning the $n\th$ datum to the $k\th$ component\\
$\widehat \phi_k$ & $ = \sum_{n=1}^N \phi_{nk}$ & effective number of assignment to the $k\th$ component in the approximate posterior
\end{tabular}
\end{table*}
\subsection{The KL-corrected bound}
The first step in deriving a collapsed bound is to select which variables should be used for parameterisation, and which should be collapsed from the problem: this can be done using a d-separation test. Given the observed variables (data) and treating the variables we wish to parameterise as observed, the collapsed variables must factorise as in the prior \citep{hensman2012fast}. Examining the graphical representation of our model in Figure \ref{fig:graphical_collapsed}, we can see that if the latent variables $\bz_{nk}$ were observed, then the model would d-separate appropriately. We shall use a variational distribution $q(\bZ)$, and introduce the variational parameters $\phi$ and the truncation level $K$ such that $q(\bZ) = \prod_{n=1}^N \prod_{k=1}^K \phi_{nk}^{\bz_{nk}}$. To ensure that $q(\bZ)$ is a valid distribution, we shall re-parameterize it through the softmax function. This will also assist in the computation of natural gradients, as we shall show. 

An important difference between our variational method and the VBEM procedure is that we do not introduce parameterisation for the distributions of collapsed variables: they will be analytically marginalised from the problem. The first step is to use Jensen's inequality to derive a lower bound on the data likelihood, conditioned on the variables that we shall be collapsing, $\bV$ and $\bF$:
\begin{equation}
	\begin{split}
		\ln p(\bY \given \bV, \bF) &= \ln \int p(\bY\given \bZ, \bF) p(\bZ\given \bV) \d \bZ\\
											    &\geq \int q(\bZ) \ln\bigg[ \frac{p(\bY \given \bZ, \bF) p(\bZ\given \bV)}{q(\bZ)}\bigg] \d \bZ \triangleq \mathcal L_1.  
	\end{split}
\end{equation}

Since $\mathcal L_1$ provides a lower bound on $\ln p(\bY \given \bV, \bF)$, we have trivially that 
$e^{\mathcal L_1}$ provides a lower bound on $p(\bY \given \bV, \bF)$. Thus we also have
\begin{equation}
	\begin{split}
		\ln p(\bY) &= \ln \int p(\bY \given \bV, \bF)p(\bF)p(\bV) \d\bV\d\bF\\
		  &\geq \ln \int e^{\mathcal L_1} p(\bF)p(\bV) \d\bV\d\bF\\
		  &\triangleq \mathcal L_\text{KL}
		\label{eq:KL_final}
	\end{split}
\end{equation}
The second integral is tractable because of the conjugacy between $e^{\mathcal {L}_1}$ and the prior. We now have a lower bound on the marginal likelihood without specifying any form for the approximate distribution of $\bF$ or $\bV$. 

\subsection{The form of the collapsed stick-breaking prior}
The integral in \eqref{eq:KL_final} separates in $\bV$ and $\bF$. The integral for $\bF$ follows easily by completing the square and using the Gaussian identity. The integral for $\bV$ is also straightforward, but reveals some relations to previous studies of collapsed stick breaking priors \citep{kurihara2007collapsed}. 

The integrals separate as follows:
\begin{equation}
	\begin{split}
		\mathcal L_\text{KL} =& \ln \int \exp\big\{\mathbb E_{q(\bZ)} \big[\ln p(\bY\given \bZ,\bF)\big]\big\}p(\bF)\d\bF\\
					     &+\ln \int \exp\big\{\mathbb E_{q(\bZ)} \big[\ln p(\bZ\given\bV)\big]\big\}p(\bV)\d\bV\\
		       &- \mathbb E_{q(\bZ)}[\ln q(\bZ)]
	\end{split}
\end{equation}
the middle term can be solved as follows. \correction{
	First, the expectation of $\ln p(\bZ\given\bV)$ is trivially
\begin{equation}
	\bbE_{q(\bZ)}\left[\ln p(\bZ\given \bV)\right] = \sumover n \sumover k \phi_{nk} \ln \pi_k = \sumover k \widehat \phi_k \ln \pi_k,
\end{equation}
where we have defined $\widehat \phi_k = \sum_{n=1}^N \phi_{nk}$. 
The stick breaking lengths $\bV$ all have beta priors with parameters $1, \alpha$. Since $v_k^0=1$ and $\Gamma(1)=1$ we have:
\begin{equation}
p(\bV) = \prod_{k=1}^\infty (1-v_k)^{\alpha-1} \Gamma(\alpha +1)\Gamma^{-1}(\alpha)
\end{equation}
Substituting these two results back into the main expression, we are left with 
\begin{equation}
\ln \int \prodover k \pi_k^{\widehat \phi_k} (1-v_k)^{\alpha-1} \Gamma(\alpha+1)\Gamma^{-1}(\alpha) \prod_{k=K+1}^{\infty} p(v_k) \d \bV
\end{equation}
Note that the part of the prior $p(\bV)$ beyond $K+1$ is trivially marginalised. Substituting the definition for $\pi_k$ and re-factoring gives
\begin{equation}
	\ln \prodover k \left\{\int v_k^{\widehat \phi_k } (1-v_k)^{\widetilde \phi_k+\alpha -1} \Gamma(\alpha+1)\Gamma^{-1}(\alpha) \d v_k\right\}
\end{equation}
where we have defined $\widetilde \phi_k = \sum_{i=k+1}^K \widehat \phi_i$.  
Finally, we recognise a series of $K$ simple beta-integrals and write
}
\begin{equation}
	\ln \prod_{k=1}^K \left(\frac{\Gamma(\widehat \phi_k + 1)\Gamma(\widetilde \phi_k +\alpha)\alpha }{\Gamma(\widehat \phi_k + \widetilde \phi_k + \alpha +1) }\right).
	\label{eq:TSB_ours}
\end{equation}

Note the similarity to the proposed collapsed stick breaking prior of Kurihara et al: 
\begin{equation}
	p_\text{TSB}(Z) = \prod_{k=1}^K \left(\frac{\Gamma(N_k + 1)\Gamma(N_{>k} +\alpha)}{\Gamma(N_{\geq k} + \alpha +1) }\right),
	\label{eq:TSB}
\end{equation}
where $N_k = \sum_{n=1}^N z_{nk}$, $N_{\geq k} = \sum_{n=1}^N \sum_{i=k+1}^K z_{nk}$. In Kurihara et~al's approach, the stick breaking lengths are marginalised from the expression {\em before} a variational approximation is made, leading to \eqref{eq:TSB}. To make this variational approximation tractable, a first order Taylor expansion of $\ln p_\text{TSB}$ is used around the point $\mathbb E_{q(\bZ)}[\bZ]$. This approximate 'marginalisation' of \eqref{eq:TSB} leads to a similar expression to \eqref{eq:TSB_ours}.

\subsection{The natural gradient in softmax}
The approximating distribution $q(\bZ)$ factorises into a $N$ multinomial distributions $q(\bz_n)$, with parameters $\bphi_n = [\phi_{n1},\ldots,\phi_{nK}]$. Since $\phi_{nk} \in [0,1]$, we use the softmax reparameterisation $\bgamma_n = [\gamma_{n1},\ldots,\gamma_{nK}]$ to avoid constrained optimization: $\phi_{nk} = \frac{e^{\gamma_{nk}}}{\sum_{j=1}^K e^{\gamma_{nj}}}$. Denoting the gradient in $\gamma$ as $\bg_n = \frac{\partial \mathcal L_\text{KL}}{\partial \bgamma_n}$, the {\em natural} gradient is given by
\begin{equation}
\widetilde \bg_n = G(\bgamma_n)^{-1}\bg_n,
\end{equation}
where $G$ is the Fisher information matrix of $q(\bz_n)$ in the parameterisation $\gamma$, which is given by $G(\bgamma_n) = \text{diag}(\bphi_n) - \bphi_n\bphi_n^\top$. This matrix is singular due to the over-parameterised nature of the softmax function, which makes computing the natural gradient problematic. \citet{kuusela2009gradient} suggests omitting the first element of $\bgamma_n$, writing $\bgamma_n' = [\gamma_{n2},\ldots,\gamma_{nK}]$, though this can be avoided as follows.

First note that inverse of $G(\bgamma_n')$ can be calculated through the
Sherman-Morrison inversion as
\begin{equation}
	G(\bgamma_n')^{-1} = \text{diag}(\bphi_n')^{-1} + {\bf 1}/(1-\sum_{k=2}^K\phi_{nk}),
\end{equation}
where ${\bf 1}$ is an appropriately sized matrix of ones. Since
$\sum_{k=2}^K\phi_{nk} = 1-\phi_{n1}$, the natural gradient is
\begin{equation}
	\widetilde \bg_n' = G(\bgamma_n')^{-1}\frac{\partial \mathcal L_\text{KL}}{\partial \bgamma_n'} = \text{diag}(\bphi_n')^{-1}\bg_n' + {\bf 1}\frac{\sum_{k=2}^Kg'_{nk}}{\phi_{n1}}
\end{equation}
We note that the symmetry of the softmax parameterisation constrains
$\sum_{k=1}^K g_{nk} = 0$, thus the gradients are $\widetilde g_{nk}'
= g_{nk}'/\phi_{nk} -g_{n1}/\phi_{n1}$. Taking a step of this length in the
variable $\bgamma'$ is equivalent to taking a step of length $\widetilde
g_{nk} = g_{nk}/\phi_{nk}$ in the variable $\bgamma$, thus the natural
gradient can be computed simply by dividing by $\phi$, with no matrix
inversions required.

Since $\bphi_{n}$ often contains many elements which are close to zero,
this division may cause numerical problems. This can be avoided by
considering the chain-rule which is used to compute the gradients with
respect to $\gamma$ from those for $\bphi$:
\begin{equation}
  \frac{\partial\mathcal L}{\partial \gamma_{nk}} = \sum_{j=1}^K \frac{\partial\mathcal L}{\partial \phi_{nj}}\frac{\partial \phi_{nj}}{\partial \gamma_{nk}} = \sum_{j=1}^K \frac{\partial\mathcal L}{\partial \phi_{nj}}(\phi_{nj}\delta_{jk} - \phi_{nj}\phi_{nk}).
\end{equation}
Dividing through by $\phi_{nk}$ obtains the following expression for the
natural gradient, which be find to be stable in computation:
\begin{equation}
  \widetilde g_{nk} = \frac{1}{\phi_{nk}} \frac{\partial\mathcal L}{\partial \gamma_{nk}} = \frac{\partial\mathcal L}{\partial \phi_{nk}} - \sum_{j=1}^K\frac{\partial\mathcal L}{\partial \phi_{nj}} \phi_{nj}. 
	\label{eq:natgrad_simple}
\end{equation}
This expression for the natural gradient is applicable for the multinomial distribution wherever the softmax parameterisation is used. 

\correction{
\subsection{Natural Gradients and VBEM updates}
We have presented the KL-corrected bound and its natural gradient for a Dirichlet process mixture of Hierarchical Gaussian processes. In the following we discuss some relations between our optimization approach and the VBEM method. 

First, it can be shown that the optimal distribution for the collapsed variables ($\bF, \bV$) factorises. Our approximate posterior thus takes the same form as the standard mean field approximation. 

Next, the mean-field bound can be shown to be exactly the same as the KL-corrected bound. If we set $q(\bZ)$ to the same distribution in each, and then update the mean-field bound with a single step of VBEM, the bounds will be equivalent \citep{hensman2012fast, sung2008latent}.  

Furthermore, the VBEM procedure is effectively a gradient method \citep{sato2001online, hoffman2012stochastic}, taking unit length steps in the natural gradient direction in each of the coordinates in turn (where coordinates correspond to the approximation to each node of the graph). An important result is that the natural gradient on the KL-corrected bound is the same as that on the mean field bound (so long as the other variables are all updated). This means that we can recover exactly the VBEM algorithm by taking unit steps in the natural gradient direction on the KL-corrected bound. 

The KL-corrected bound has then brought about the following advantages: it is simpler to derive (there are fewer variables to deal with), and it has a lower-dimensional space for optimization. Surprisingly, this more compact representation {\em does not} complicate the optimization: natural gradient steps on the KL-corrected bound have the same effect as a full set of steps on the mean-field bound (or a full round of updates). 

Perhaps the most important advantage of the KL-corrected bound is that it enables {\em conjugate gradient descent} to be preformed easily. One only needs to consider the conjugate computations (as presented by \citet{honkela2010approximate}) in a small number of variables. If the conjugate gradient step fails to improve the bound, then reverting to a unit step in the natural gradient direction will recover the VBEM update again, and the conjugate part of the algorithm can be 'restarted'. 
}

\subsection{A merge-split procedure}
A merge-split approach has been suggested for mixture models using EM \citep{ueda2000split} and in MCMC \citep{jain2004split}. In this approach, the current solution is re-initialised by either re-defining two clusters as one, or one cluster as two new. The collapsed nature of the KL-corrected bound is particularly helpful in performing merge-split, since we only have the parameters $\phi$ (or equivalently $\gamma$) to deal with. Since we have a lower bound on the marginal likelihood, we also have a  natural method for accepting proposed moves, depending on whether they increase the bound. 

To perform a split, we select a cluster component $k$ and find the data which are currently associated by examining $\phi_{nk}$. We increase the truncation parameter $K$, adding a new cluster, and move half of the probabilistic mass for $\phi_{nk}$ to this new cluster $\phi_{nK}$. After optimising to convergence, we accept the move if the bound increases. We found empirically that merge procedures were not necessary: that optimization naturally managed to merge clusters appropriately. We simply removed empty clusters as appropriate. We also make use of the re-ordering move \citep{kurihara2007collapsed} which re-order the solution so that the largest cluster is first: this increases the bound under the DP prior. 

\correction{
We note that the collapsed parameterisation of the model makes these procedures
simple to implement: with only a $N\times K$ matrix containing $\phi_{nk}$ to
deal with, columns (corresponding to clusters) can be deleted, added, moved or
adjusted arbitrarily, so long as the bound on the log-likelihood increases. 
}

\section{Experiments}
We present the application of our model and variational inference procedure to three data sets. \correction{
	In all the experiments, we initialised the allocation of clusters randomly. The effect of the covariance function hyper-parameters can play quite a strong role in the results. We initialised using the following rules of thumb: length-scales were initialised to half of the span of the input data; the top-level variance (cluster variance) was set to account for 60\% of the signal variance; the hierarchical variance was set to 30\% and noise was set to account for 10\% of the data variance. Optimization of the hyper-parameters (against the lower bound on the marginal likelihood) was interleaved with the variational optimization. Unless otherwise stated, the Dirichlet process concentration parameter was fixed to 1. }

\subsection{Data sets and models}
\label{par:dros_dev}
\subsubsection{Synthetic data}
To demonstrate our model, we generated a synthetic data set as follows. We selected 12 time points randomly in the region $(0,1)$, and defined $10$ clusters by evaluating the sine function with uniformly randomised phase and randomised frequency around $2\pi$. We randomly selected $N_k$ data per cluster in the interval $(20,30)$, and for each datum in each cluster selected a correlated offset from the mean for each cluster using further randomised sine functions, and added a small amount of i.i.d. noise. The data are illustrated in Figure \ref{fig:synthetic}. 

\subsubsection{Drosophila development}
We present results of clustering data from \citet{kalinka2010gene}. In this paper, the gene expression of six species of {\em Drosophila} was presented, measured at two hour intervals during embryonic development. The data contains a natural structure: aside from structure across species, the experiments was performed in replicate. Pools of embryos were used, with measurements taken every two hours from each pool. Not all of the pools were measured at all time points. We use a hierarchical GP to model this replicate structure, accounting for correlated differences between replicates. We use a further level of the hierarchy for clustering the genes. Using a method similar to \citet{kalaitzis2011simple} to eliminate silent genes, we selected 1000 genes for clustering. 

\subsubsection{Periodic clustering}
\begin{figure*}
	\centering
	\includegraphics[width=\textwidth]{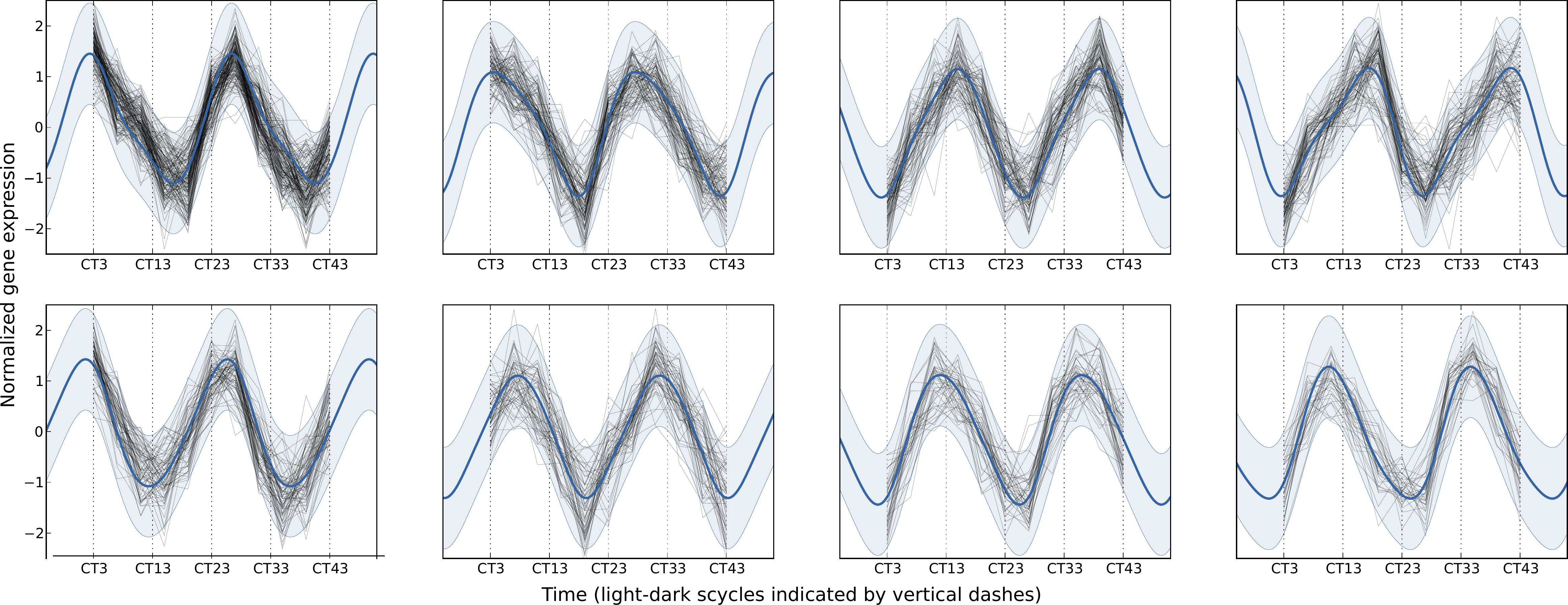}
	\caption{\label{fig:circadian} Eight of the ten Clusters found in the mouse cartilage data using periodic GPs for clustering. In this case, the only information shared between genes in a group must be captured by a periodic GP. We are fortunate enough to be given the period of the rhythm in advance: it is 24hrs as enforced by the light-dark cycle. Although there are other effects in the data, in this case we do not wish to use them in clustering: they are thus modelled on a gene-by-gene basis as a RBF (and i.i.d. noise) GP. }
\end{figure*}

An advantage of using GP models for clustering is that we can incorporate prior information into the model. In \citet{gossan2013circadian} gene expression in mouse cartilage was measured at four hour intervals in duplicate, following a 12h-12h light-dark cycle. 304 genes corresponding to  circadian rhythms were identified by fitting sine functions to the data. Here we propose a clustering model for these circadian genes. We use a periodic covariance function based on a projection of the Matern covariance \citep{durrande2013gaussian}, drawing functions from the DP-GP construct which are periodic (not necessarily sinusoidal) in nature. The next layer of our hierarchical structure uses the standard RBF covariance with i.i.d. noise. In this model, the genes are able to share only a {\em periodic} component: deviation from this periodicity is accounted for on a gene-by-gene basis. Clusters inferred by our model are shown in Figure \ref{fig:circadian}. Further analysis of the discovered cluster structure showed that it reflected known groupings of established clock genes as well as providing insights into the regulation of cartilage specific genes by the circadian clock. For more details, see \citet{gossan2013circadian}. 

\subsection{The importance of structure}
\begin{figure*}
	\centering
	\includegraphics[width=\textwidth]{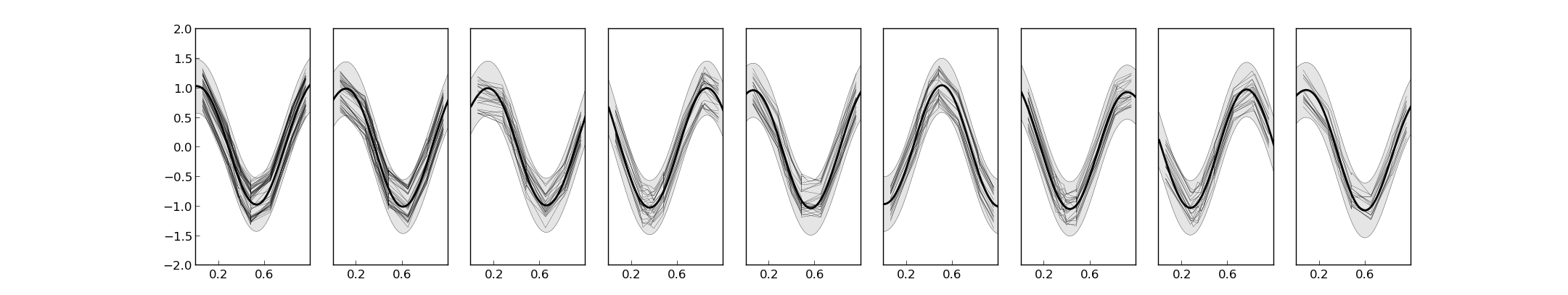}
	\caption{\label{fig:synthetic} The synthetic data set, shown in the clustering formation inferred by the hierarchical model. Each pane represents one cluster, and data assigned to that cluster are represented as thin lines joining the observations. Posterior means and two standard deviations of $f$ are shown as solid lines and shaded areas. We note that this Figure omits some additional structure: each datum is modelled as a GP (not shown) whose prior mean is that common to the cluster.}
\end{figure*}

We tested our model on the synthetic data, and compared to GP-DP construct {\em without} a hierarchical structure, using only i.i.d. noise to model the difference in clustered signals, and a Dirichlet process Gaussian mixture model, which attempts to infer all the covariance structure in the data. For comparison, we set the concentration parameter $\alpha$ to give the correct number of clusters {\em a priori}, and inferred using the same scheme for each. 
\begin{figure}
	\centering
	\includegraphics[width=\linewidth]{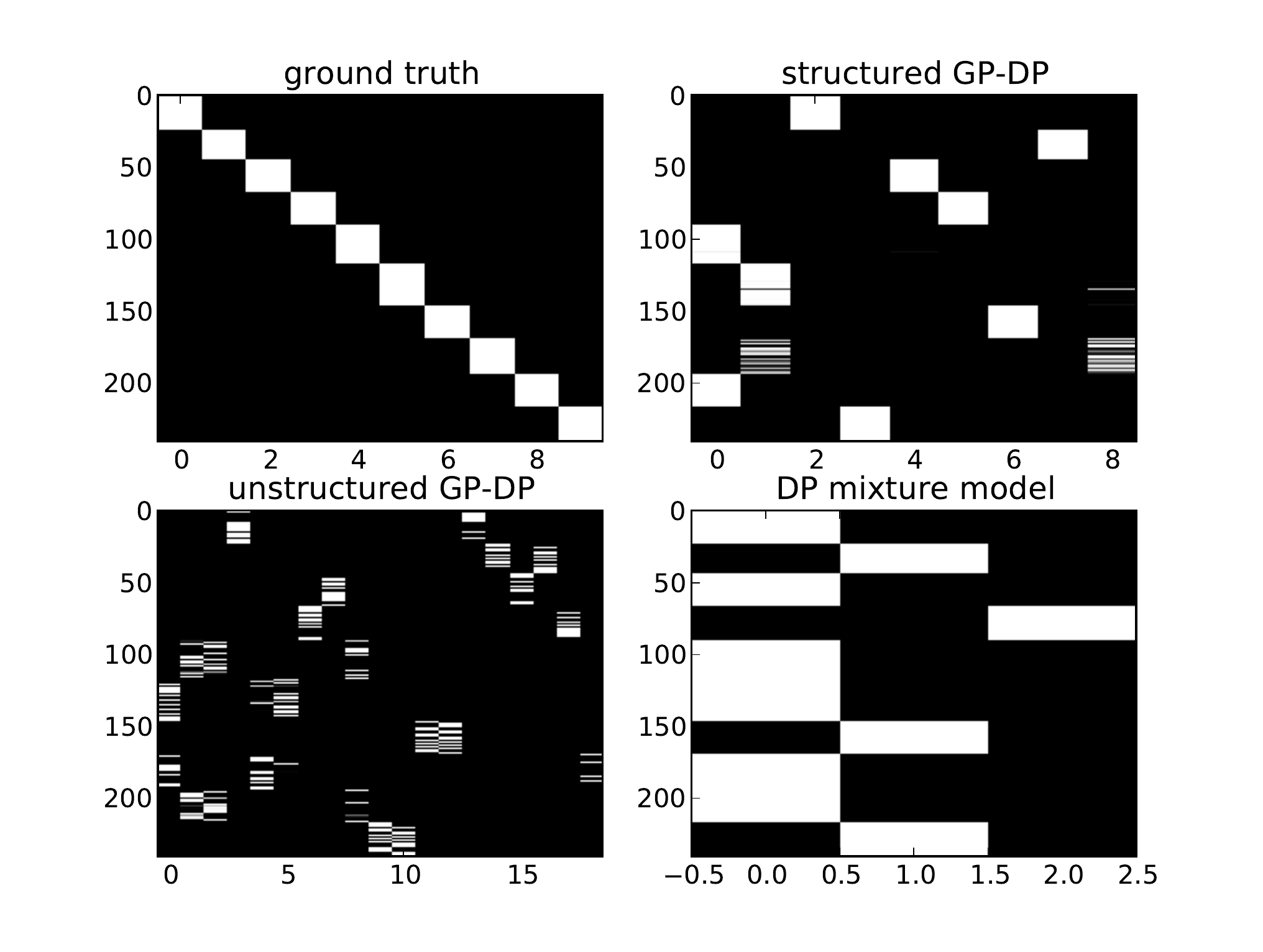}
	\caption{\label{fig:hintons} Cluster allocation diagrams for the synthetic data. In each, the data are indexed vertically and the clusters are index horizontally: white square indicates allocation of a datum to a cluster, grey squares represent uncertain allocations. The hierarchical GP model finds most of the correct structure, with some confusion in the first and second(eighth) cluster. The non-structured GP-DP fails to correctly account for structure in the data, and uses too many clusters to model the variance. The DP Gaussian mixture model is unable to discern the clusters correctly. }
	\end{figure}
Figure \ref{fig:hintons} shows the ground truth cluster allocation and the inferred cluster allocation for each. From the Figure we see that our model has correctly inferred most of the correct structure: only two true clusters are confused. The GP-DP construct without a structured model is poor in finding the clusters: since it cannot account for correlations amongst signals using a noise model, it attempts to introduce extra components into the model to account for variance. The Gaussian mixture model also fails to infer the correct structure: without prior knowledge of signal correlations, it is unable to separate the groups. The clusters inferred by our model are shown in Figure \ref{fig:synthetic}. 

To optimize the parameters of the covariance functions, we interleave standard our Riemannian method with standard optimization of the covariance functions parameters, keeping the variational distributions fixed. Using the synthetic data, we set up a simple trial to examine the sensitivity of the method to the initial conditions of the covariance function parameters. We created 20 initial conditions, drawing parameters values from the standard log-normal distribution, and optimized the models using the Riemannian procedure interleaved with standard conjugate-gradient optimization of the parameters and a merge-split routine. In 16 of the 20 cases, the optimal structure (as shown in Figure \ref{fig:hintons}) was discovered; in the remaining 4 cases, two of the clusters were conflated, which was reflected by a lower bound on the marginal likelihood, whilst the remaining structure was inferred correctly. In all cases, the lengthscales and variances of the covariance functions were estimated correctly, to with 2 decimal places of the best solution. 
\begin{figure*}
	\centering
	\includegraphics[width=\textwidth]{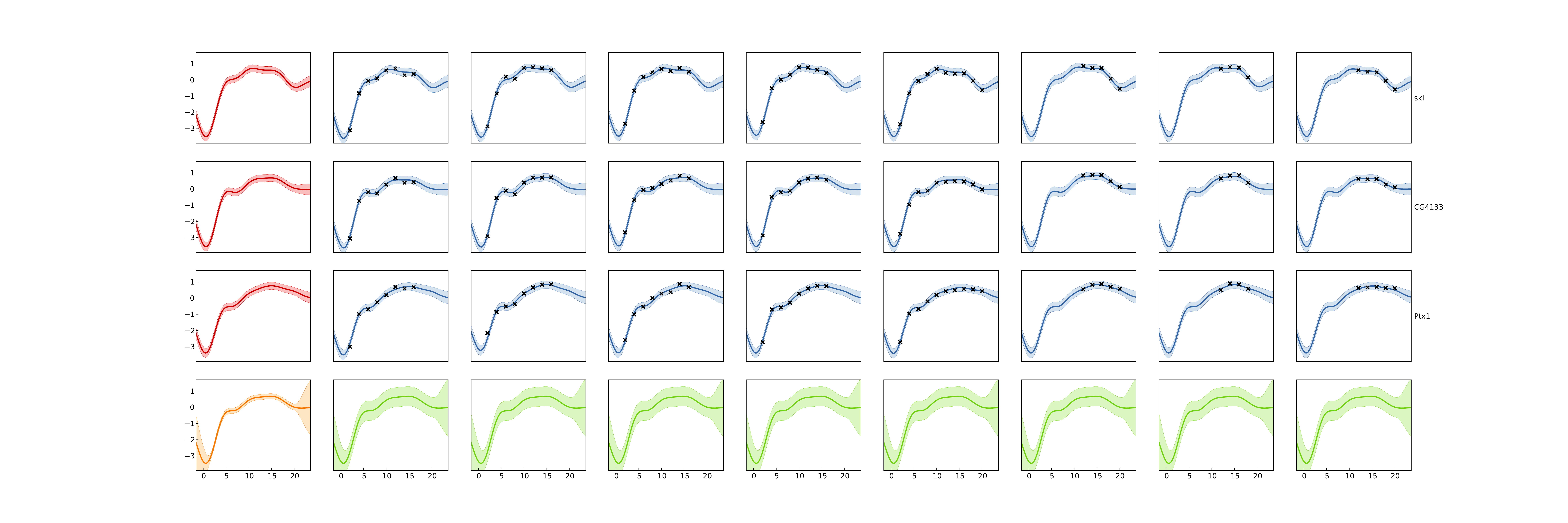}
	\caption{\label{fig:dros_structure} An example of the structure inferred within a single cluster of the {\em Drosophila} data set. The function $f$ which governs the behaviour of the cluster is shown in the bottom left panel. Each row represents a single gene in the cluster, with the left-most pane representing the inferred function for that gene, and subsequent panes representing the inferred function for individual replicates. The bottom row shows the predictive distribution for a hypothetic extra gene in the cluster.}
\end{figure*}

The results of clustering the Drosophila data are shown in the supplementary material. An example of the structure inferred in a single cluster is illustrated in Figure \ref{fig:dros_structure}.  The use of our hierarchical model allows us to find biologically 
meaningful clusterings that may not be possible without the structure.
For example, the first three clusters appear to
be very similar: signals in the second cluster rise only slightly
faster than in the first. Using the online DAVID tool, we found that
the cellular component gene ontology (CCGO) terms were enriched
differently in each. In the first cluster, the top CCGO terms were
{\em extracelular matrix}, {\em extracelluar region part} and {\em
proteinaceous extracellular matrix}, with p-value belows 0.005. In the
second cluster, the  top CCGO terms were {\em plasma membrane part}
and {\em cell junction}, with similar p-values. The third cluster,
whose signals arrive slightly later still, was also enriched for genes
{\em intrinsic to membrane}, but also showed enrichment for {\em cell
adhesion} and {\em biological adhesion}.

\correction{
	For comparative purposes, we also applied our code to clustering the data {\em without} the hierarchical structure. This is similar in spirit to that proposed in \citet{dunson2010nonparametric}, though we maintain our variational framework. In this model, all variation from the cluster mean is modelled as independent Gaussian noise. The result is that many more clusters have to be used to model the data. Noise in the measurement process is not simply i.i.d. as this model enforces: varying sensitivities of the microarray system as well as true biological variation in the genes mean that those genes activated by similar pathways -- thus having similar temporal patterns -- will vary in a correlated manner. We summarise the results of the two models in Table \ref{tab:dros_compare}.
\begin{table}[!h]
\caption{Comparison of the hierarchical and non-hierarchical methods on the {\em Drosophila} data}
	\centering
	\begin{tabular}{|c|c|p{3cm}|}
		\hline
		& Hierarchical & Non-hierarchical\newline \citep[similar to][]{dunson2010nonparametric}\\
		\hline
		N. clusters &52 &245\\
		$\mathcal L_\textsc{KL}$ &4256.2&-58.9\\
		Signal variance & 1.26 & 1.41\\
		Noise variance & 0.03 & 0.04\\
		hierarchical variance & 0.06& n/a\\
		\hline
	\end{tabular}
	\label{tab:dros_compare}
\end{table}

We first note that the lower bound on the marginal likelihood is dramatically higher for the hierarchical model.  Even accounting for the few extra parameters required, the difference is extremely significant. From the table we see that the hierarchical model uses a smaller noise level to model the data, and uses twice the variance of the noise to model hierarchical structure (including both replicate and gene-wise variance).  Plots of the discovered clusters can be found in the supplementary material -- it is clear that the non-hierarchical version discovers many small clusters with very similar profiles.  
}

\subsection{Efficiency of the inference procedure}
To perform inference, we used the VB procedure described with a merge-split, and used a gradient-based method to find point-estimates of the kernel parameters, based on maximising $\mathcal L_\text{KL}$.

Subsequently, to compare our Riemann procedure with VBEM and to test the effectiveness of the merge-split approach, we set the kernel parameters to sensible values found using several optimization runs. 

\begin{figure}
	\centering
	\includegraphics[width=0.8\linewidth]{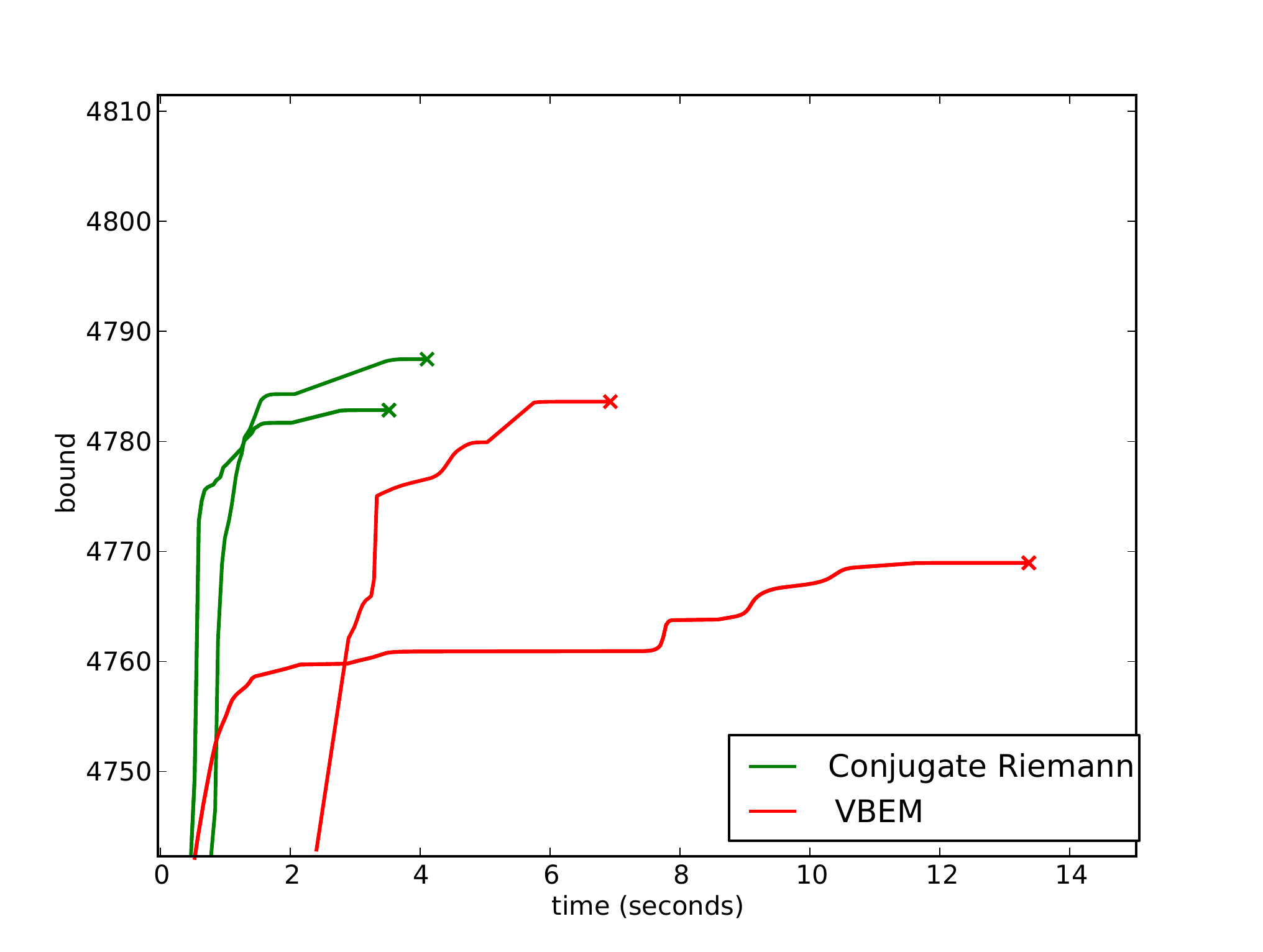}
	\caption{\label{fig:example_convergence} Convergence of our method on the {\em Drosophila} dataset, using two random restarts. The same restarts were applied to both methods. The conjugate Riemann method uses Hestenes-Steifel conjugacy on the manifold, whist the VBEM procedure is effectively steepest descent. Both types of optimization show that there are plateaus in the objective function: the conjugate method quickly escapes these, whilst VBEM can only move according to the local gradient, and becomes stuck. }
\end{figure}

Previous uses of the collapsed VB method have reported that it finds superior solutions \citep{sung2008latent}.  We empirically found that this occurred in our algorithm: the VBEM procedure, which is a steepest ascent method on the Riemann manifold, becomes stuck on plateaus where there is little gradient. This is illustrated by Figure \ref{fig:example_convergence}: both the conjugate Riemann method and VBEM pause at the same levels of likelihood, but the nature of our algorithm allows it to pass through this solution, whilst the VBEM algorithm is stuck.

To monitor the effects of this on the ability of the algorithm to find a good solution, we ran 200 restarts for each of our data sets, using the same initial conditions for each without using the merge-split approach. We then monitored, over all the restarts, how many times the algorithm came to  good solution (which we defined as being within 10 nats of the best-found solution). We divided the total time taken (or iterations used) for all 200 restarts by the number of runs which found such a solution. This statistic then asses how well the algorithms perform in not only speeding up convergence but also escaping the plateaus as discussed.  The results are shown in Table \ref{tab:results}. 

\begin{table*}
	\caption{Timing of clustering each of the datasets using VBEM and the Riemann approach.\label{tab:results}}
	\centering
	\begin{tabular}{|c|c|c|c|c|}
		Dataset (N. genes) & VBEM iters. &  Riemann iters. & VBEM (s)&  Riemann (s)\\
		\hline
		Synthetic (241) &304&234&1.16&0.90\\
		Drosophila (600) & 680& 381&153&88\\
		mouse cartilage (896) & 232 & 102 & 34 & 14\\
	\end{tabular}
\end{table*}

The effect of expedited convergence using the Riemann approach is amplified when optimising the hyperparameters and using the merge-split method. In practise, it is necessary to run the variational optimization many times, interleaved between merge-split trials and optimization of the hyper-parameters. Since the Riemann procedure often finds better local solutions for the variational parameters, it does not need to use as many split-procedures to find a good global optimum. When optimising the hyper-parameters interleaved with the VB parameters, the Riemann procedure is also particularly effective, finding local solutions more quickly. 

\section{Summary \& Discussion}  
We have presented a method for clustering of structured time series. The method is based on hierarchical Gaussian process. This simple idea allows us to combine related time series groups in a natural way. We introduced a clustering model which allows us to infer the groupings, modelling further structure within sub-groups. 

Inspired by gene expression time series, we applied our model to three datasets. We showed how the Gaussian process methodology allows us to incorporate knowledge of the problem into the model such as periodicity of the shared time-series function. The model has many applications in clustering time series, and we are currently exploring the application to motion capture data. 

We performed inference in the model using a recent modification of variational Bayes. This not only provided a speed improvement, but also allowed for extremely simple implementations of a merge-split approach.  We related the collapsed expression to that used in collapsed variational Bayes, and showed how to compute the natural gradient for a set of softmax-parameterised variables, a derivation which has wide application in clustering models. 

A {\em python} implementation of the algorithm and code for running all the experiments can be found on our website at {http://staffwww.dcs.shef.ac.uk/people/J.Hensman/}\, .


\bibliographystyle{abbrvnat}

\bibliography{references}

\end{document}